%% file: main.tex
\documentclass[sigconf]{acmart}
\AtBeginDocument{%
  \providecommand\BibTeX{{%
    \normalfont B\kern-0.5em{\scshape i\kern-0.25em b}\kern-0.8em\TeX}}}

\acmPrice{15.00}

\copyrightyear{2022}
\acmYear{2022}
\setcopyright{rightsretained}
\acmConference[KDD '22]{Proceedings of the 28th ACM SIGKDD Conference on Knowledge Discovery and Data Mining}{August 14--18, 2022}{Washington, DC, USA}
\acmBooktitle{Proceedings of the 28th ACM SIGKDD Conference on Knowledge Discovery and Data Mining (KDD '22), August 14--18, 2022, Washington, DC, USA}
\acmISBN{978-1-4503-9385-0/22/08}
\acmDOI{10.1145/3534678.3539208}
\usepackage{etoolbox}
\makeatletter
\patchcmd{\maketitle}{\@copyrightpermission}{
   \begin{minipage}{0.3\columnwidth}
     \href{https://creativecommons.org/licenses/by/4.0/}{\includegraphics[width=0.90\textwidth]{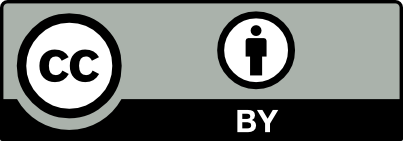}}
   \end{minipage}\hfill
   \begin{minipage}{0.7\columnwidth}
     \href{https://creativecommons.org/licenses/by/4.0/}{This work is licensed under a Creative Commons Attribution International 4.0 License.}
   \end{minipage}
 
   \vspace{5pt}
}{}{}

\makeatother

\usepackage{svg}
\usepackage{amsmath}
\usepackage{caption}
\usepackage{subcaption}

\begin{document}
\newcommand{\rongzhi}[1]{\textcolor{blue}{[Rongzhi: #1]}}
\newcommand{\xc}[1]{\textcolor{cyan}{[Xiquan: #1]}}
\newcommand{\ours}{\textsc{AMRule}}
\synctex=1
\settopmatter{printacmref=true}
\newcommand{\zc}[1]{\textcolor{blue}{[#1 - Chao]}}

\title{Adaptive Multi-view Rule Discovery for Weakly-Supervised Compatible Products Prediction}

\author{Rongzhi Zhang}
\affiliation{%
  \institution{Georgia Institute of Technology}
  \city{Atlanta}
  \state{GA}
  \country{USA}}
\email{rongzhi.zhang@gatech.edu}

\author{Rebecca West}
\affiliation{%
  \institution{The Home Depot}
  \city{Atlanta}
  \state{GA}
  \country{USA}}
\email{rebecca_west@homedepot.com}

\author{Xiquan Cui}
\affiliation{%
  \institution{The Home Depot}
  \city{Atlanta}
  \state{GA}
  \country{USA}}
\email{xiquan_cui@homedepot.com}

\author{Chao Zhang}
\affiliation{%
  \institution{Georgia Institute of Technology}
  \city{Atlanta}
  \state{GA}
  \country{USA}}
\email{chaozhang@gatech.edu}
\newcommand{\yue}[1]{{\text{\textcolor{red}{[Yue: #1]}}}}
\renewcommand{\shortauthors}{Rongzhi Zhang et al.}

\begin{abstract}





  On e-commerce platforms, predicting if two products are compatible with each
  other is an important functionality to achieve trustworthy product
  recommendation and search experience for consumers. However, accurately
  predicting product compatibility is difficult due to the heterogeneous
  product data and the lack of manually curated training data. We study the
  problem of discovering effective labeling rules that can enable
  weakly-supervised product compatibility prediction. We develop \ours, a
  multi-view rule discovery framework that can (1) adaptively and iteratively
  discover novel rulers that can complement the current weakly-supervised
  model to improve compatibility prediction; (2) discover interpretable rules
  from both structured attribute tables and unstructured product descriptions.
  \ours~adaptively discovers labeling rules from large-error instances via a
  boosting-style strategy, the high-quality rules can remedy the current
  model's weak spots and refine the model iteratively. For rule discovery from
  structured product attributes, we generate composable high-order rules from
  decision trees; and for rule discovery from unstructured product
  descriptions, we generate prompt-based rules from a pre-trained language
  model. Experiments on 4 real-world datasets show that \ours~outperforms the
  baselines by $5.98\%$ on average and improves rule quality and rule proposal
  efficiency.

\end{abstract}





\maketitle

\input{01-intro}
\input{02-related}

\input{03-prelim}
\input{04-method}
\input{05-exp}
\input{06-conclusion}
\begin{acks}
  This work is performed under the generous support of a gift fund from The Home Depot. Chao Zhang was supported in part by NSF IIS-2008334, IIS-2106961, CAREER IIS-2144338.
\end{acks}


\bibliographystyle{ACM-Reference-Format}
\bibliography{sample-base}

\appendix

\end{document}

%% file: 01-intro.tex
\section{Introduction}
\begin{figure}
    \centering
    \includegraphics[scale=0.2]{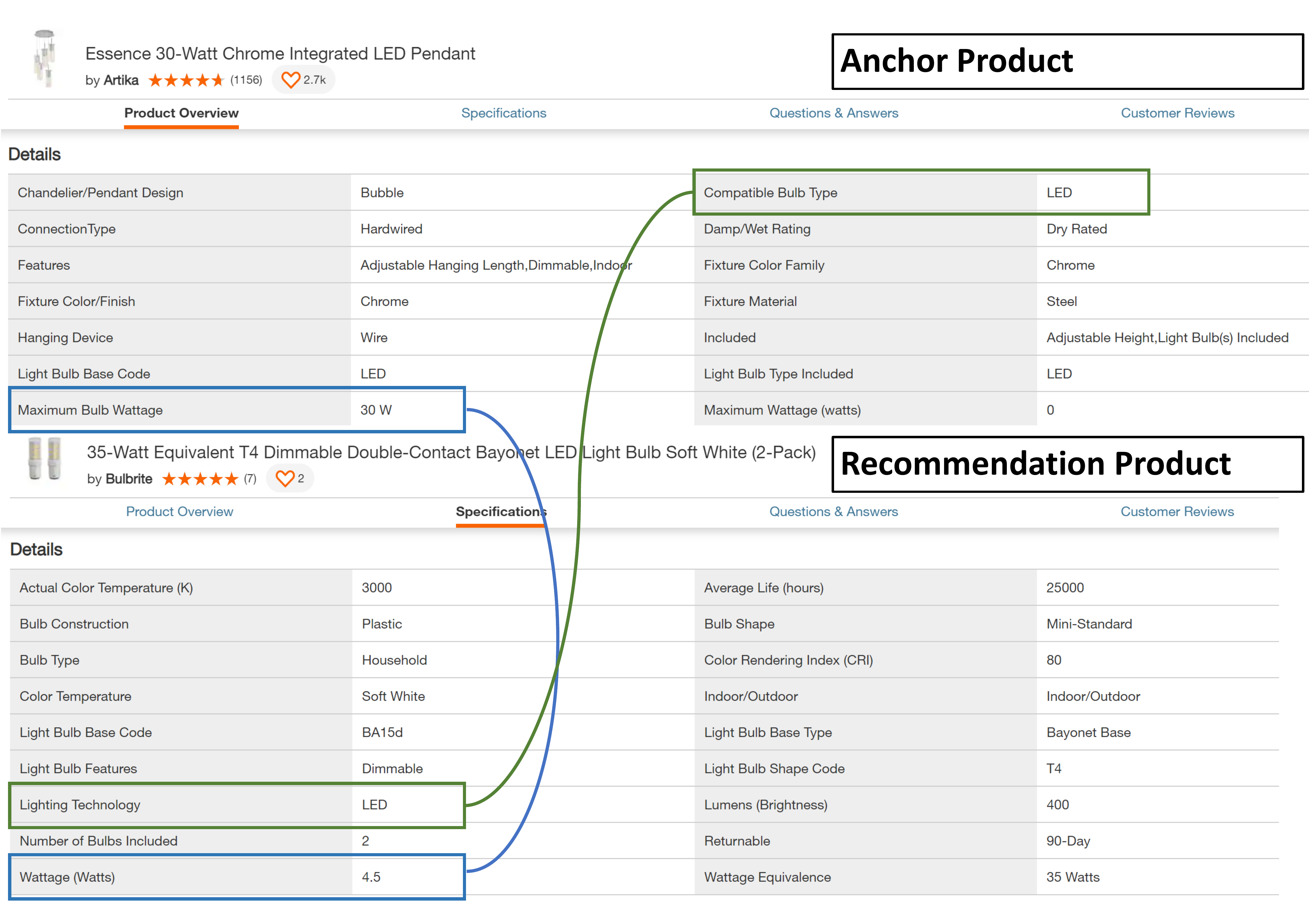}
    \vspace{-5mm}
    \caption{An example of compatible products on homedepot.com. The product attributes highlighted by the box with the same color indicate the compatibility.}
    \label{fig:comp_example}
\end{figure}

The prevalence of bundle items on e-commerce and content platforms makes identifying compatible products become an important task.
The 
product compatibility prediction problem aims to predict if two products are
functionally compatible with each other. 
Figure \ref{fig:comp_example} shows a concrete example on The Home Depot
online platform \footnote{https://www.homedepot.com}. To ensure a chandelier
and a bulb are compatible, it is required that their lighting technologies are
matched and the wattage is within-range---all such information needs to be
inferred from their product data. Accurate compatible products recommendation
is critical to e-commerce platforms, as this functionality is key to ensuring
trustworthy and reliable customer experiences in product search
recommendation.


Despite the importance of this task, accurate product compatibility prediction
is challenging due to two key reasons: 1) the \textit{lack of manually curated
  data to train an accurate classifier}; and 2) the \textit{heterogeneity of
  product data. } First, we typically do not have clean labeled data for
learning a product compatibility predictor. Manually labeling compatible
products is very time-consuming, because the annotators need to inspect the
high-dimension product attributes and compare multiple entries between anchor
products and recommendation products. Although one may resort to user
behaviors such as co-purchase records as pseudo-positive labels, these data
could be noisy because co-purchase behavior does not always reflect
compatibility of two items. Second, the representation of products includes
both structured product attributes and unstructured product descriptions,
where product attributes have high cardinality, and the product description is
not an ideal source for mining the buried compatibility. In practice, the
compatibility rules are written by domain experts in The Home Depot. Although
more efficient than labeling individual data points, the process of developing
rules still requires intensive human labor, and most of the human-written
rules just consider one product attribute while ignoring high-order
association of product attributes.
 
Compatible product prediction is mainly related to two lines of work in
literature. One line of work is weakly-supervised learning (WSL), which has shown
promising results in addressing label scarcity in various tasks including
\emph{text classification}
\cite{awasthi2020learning,meng2020text,yu2021fine,zhou2020nero}, \emph{dialogue
  systems}~\cite{MallinarSUGGHLZ19}, and \emph{fraud
  detection}~\cite{zhang2021fraud}. However, the performance
of WSL approaches is often hindered by the low quality of the \emph{initial rules}
and the \emph{static learning process}~\cite{zhang2022prboost}. 
For the initial rules, either the human-written rules \cite{ratner2017snorkel, hancock2018training}
requires tedious and time-consuming creation, or the automatically discovered rules
\cite{varma2018snuba} are restricted to frequent patterns and predefined types
which are inadequate for training an accurate model. 
For the static learning process used in ~\cite{ratner2017snorkel,zhou2020nero,meng2020text}, 
the performance of these methods largely depends on the quality of the initial weak sources.
When the quality of the weak labels is poor, the incorrect weak labels can result in error propagation and deteriorate the model's performance. 
To improve the quality of the rules, interactive rule discovery~\cite{boecking2020interactive, galhotra2021adaptive,hsieh2022nemo}  has been explored to iteratively refine the rule set using human guidance. However, \cite{boecking2020interactive} only discover simple rules with a repetitive structure given by a pre-defined rule family, and ~\cite{galhotra2021adaptive} uses an enumerating-pruning approach to search rules, which is inapplicable for large corpus.

The second line is pattern-based classification \cite{chicco2004load,shang2016dpclass,cheng2007discriminative}. Classical
approaches construct a pattern pool based on pattern frequency and then select
the most discriminative patterns via heuristics \cite{cheng2007discriminative,fan2008direct}. However, the large pattern pool makes the computation
inefficient, and the number of selected patterns limits the interpretability.
To tackle the drawbacks, \citet{shang2016dpclass} proposes to generate prefix
paths in tree-based models and further compress the number of patterns by
selecting the most effective ones. \citet{liu2021controlburn} prune
unnecessary features from tree ensembles using a weighted LASSO-based method.
Although these methods improve the interpretability and efficiency, they have
not explored the weakly supervised settings, under which we aim to discover
compatibility rules and perform compatible products prediction.
 
We propose an adaptive multi-view rule discovery framework \ours~ for weakly-supervised compatible products prediction. Our approach overcomes the aforementioned challenges
with the following key designs:
\begin{itemize}
    \item To address the label scarcity issue of product compatibility
      prediction, we leverage user behavior data as the supervision source to generate weakly labeled instances. Specifically, we use co-purchase data to generate positive labels and randomly sample product pairs from the anchor categories and the recommendation categories as negative labels.
    \item To avoid error propagation caused by the imperfect initial weak source, we design a boosting strategy to iteratively target rule discovery on difficult instances. In each iteration, we reweigh data by boosting error to enforce the rule discovery module to focus on larger-error instances. We prevent enumerating massive possible rules and post-filtering for novel rules, but directly propose rules based on large-error instances to provide complementary information to the current model.
    \item To handle the complex input of products, we propose a multi-view rule discovery framework, where we integrate two rule representations based on structured and unstructured data. Specifically, we design a decision tree-based rule generation method where the key product attributes are selected to form a composable rule. As the product attributes are high-cardinality and not consistent between product pairs, we further leverage product description to generate a fallback rule, where the rules are generated by prompting pre-trained language models (PLMs) to capture the hidden semantics in the unstructured text-format description. 
\end{itemize}
 
We conduct extensive experiments on the Home Depot real datasets across four
categories of products. The results show that 1) the proposed method
outperforms all the baselines, including static approaches and iterative
approaches; 2) The boosting strategy is effective in discovering novel rules
by targeting hard instances iteratively; 3) The multi-view rule discovery
design is effective, and the rules generated from structured attributes and
unstructured descriptions can complement each other; 4) Our rule discovery
pipeline improves the interpretability and annotation efficiency.
 
 Our main contributions can be summarized as follows: 1) a user behavior-based approach for weakly supervised compatible product prediction; 2) a multi-view rule discovery design integrating structured and unstructured data representation; 3) an iterative and adaptive rule discovery framework via model boosting; 4) comprehensive experiments on  real-world datasets verifying the efficacy of our method.


%% file: 02-related.tex
\section{Related Work}

\paragraph{Weakly Supervised Learning} 
The weak supervision learning (WSL) frameworks provided a solution to truly reduce the efforts of annotation~\cite{ratner2017snorkel}.  
Specifically, in WSL, users encode \emph{weak supervision sources}, e.g., heuristics, knowledge bases, and pre-trained models, in the form of \emph{labeling functions (LFs)}, which are user-defined programs that each provide labels for some subset of the data, collectively generating a large set of training labels~\cite{zhang2021wrench}.
Weak supervision methods have been widely applied in various domains including text classification~\cite{awasthi2020learning,meng2020text,zhang2022understanding,yu2020steam, zhang-etal-2020-seqmix}, dialogue systems~\cite{MallinarSUGGHLZ19}. 

The success of the WSL paradigm has attracted much attention from the industry. For example,
\citet{tang2019msuru} deploy an image understanding system in Facebook
Marketplace, where the user interactions in search logs serve as weak supervision sources for linking text-format queries to images and enables the curation of a large-scale image understanding task.
However, they focus on how to leverage the weak signals in the user
interaction data without denoising the weak supervision source.
\citet{xiao2019weakly} study weakly supervised query rewriting and semantic
matching for e-commerce platforms, they also leverage search logs to build an
unlabeled dataset. They propose a co-training framework where two models for
different tasks generate weak labels for each other to boost their performance
simultaneously. \citet{zhan2021product1m} explore multi-modal product
retrieval on e-commerce dataset. They generate pseudo-labels for the
image-text pairs by a copy-and-paste data augmentation approach, and overcome
the fuzzy correspondence between modalities via contrastive learning. Despite
the successful deployment of these methods, they are tailored for specific
applications and thus cannot be transformed to the compatibility prediction
task. Their end-to-end learning paradigm also sacrifices the interpretability
of the framework, but in our task, we aim for model performance improvement
and also interpretability of the weakly labeled data.

\paragraph{Rule Discovery} 
Most work of WSL relies on the human-written rules as weak sources, but manually designing the rules can be tedious and time-consuming. To alleviate human efforts in the rule discovery process,
a few works attempt to discover rules from data.
For example, Snuba~\cite{varma2018snuba} automatically generates rules using a small labeled dataset to create weak labels for another large, unlabeled dataset; 
TagRuler ~\cite{choi2021tagruler} also proposes an automatic creation of labeling functions based on the human demonstration  for span-level annotation tasks. 
However, these methods constrain the rules to task-specific types and the rule set cannot be modified once proposed. Instead, we integrate multiple views of rules for the heterogeneous data and design an adaptive rule discovery framework, which iteratively complements the existing rule set.
Two works have studied \emph{interactive
WSL} \cite{boecking2020interactive,galhotra2021adaptive}
as in our problem. However, they either use simple $n$-gram-based rules
\cite{boecking2020interactive} that are not applicable to our problem, or suffer from a huge searching space for context-free grammar rules \cite{galhotra2021adaptive}. 
Although \cite{zhang2022prboost} also leverage prompt for rule generation, it mainly focuses on textual data, and cannot be well applied to our scenarios where both the structured and unstructured features exist. 
In contrast, our proposed multi-view rules consider both structured and unstructured information of products, and the adaptive rule discovery process avoids enumerating all possible rules and performing post-filtering for novel rules.

\paragraph{Prompting Methods for Pre-trained Language Models} 
The rule discovery component in our work is also related to prompting methods for PLMs. 
The prevailing prompting methods modify the original input text into a string with unfilled slots, from which the final prediction can be derived~\cite{brown2020language,liu2021pre}.
It enables PLMs to adapt to new scenarios with few or no labeled data~\cite{hu2021knowledgeable,han2021ptr, lester2021power,chen2021adaprompt, li2021prefix}. Recent works explore learning implicit prompt embedding via tuning the original prompt templates~\cite{gao2021making, liu2021p,liu2021gpt, vu2021spot}. Instead of tuning prompts to achieve optimal performance for the original task, we leverage prompts and PLMs to assist the multi-view rule discovery.  Moreover, none of
these works studied prompting to generate weak labels for WSL as we have.

%% file: 03-prelim.tex
\section{Preliminaries}
Compatible products prediction underpins the recommendation of sale bundles on e-commerce websites. 
We first introduce the representation of products here. 
Given a product category $\mathcal{X}_a$ and a specific product $\boldsymbol{x}_a \in \mathcal{X}_a$, the representation of $\boldsymbol{x}_a$ includes two parts: the \emph{structured product attributes} $\mathbf{w}_a = [w_1, w_2, \cdots, w_n]$, and the \emph{unstructured product description} $\mathbf{\tau}_a$. For a product category $\mathcal{X}_a$, there is an attribute set $\mathcal{W}_a$, such that $\forall \boldsymbol{x}_a \in \mathcal{X}_a, \mathbf{w}_a \in \mathcal{W}_a$. 

Besides, we leverage the item-level co-purchase data as the weak supervision source. Each entry in the item-level co-purchase data is a tuple $(\boldsymbol{x}_a, \boldsymbol{x}_b, \nu)$, where $\nu$ is the co-purchase times. Note that the records in co-purchase data are not necessarily compatible products, users could purchase two items together for other reasons rather than compatibility.
Now we give definitions of the task:

\begin{definition}[Compatible products prediction]
    For two mutually exclusive product corpus $\mathcal{X}_a$ and  $\mathcal{X}_b$, given an anchor product $\boldsymbol{x}_a \in \mathcal{X}_a$ and a recommendation product $\boldsymbol{x}_b \in \mathcal{X}_b$,  we aim to predict the label $y \in \{1, -1\}$ indicating whether the product pair can be 
    functionally
used together without conflict.
\end{definition}

Because the dataset curation involves user behavior data as weak supervision signals, it is necessary to generate more high-quality labeled data via a reliable source to suppress the noise in the initial dataset. For the compatible product prediction task, compatibility rules (labeling rules) are a reasonable way to capture the relevance between the product attributes of the product pairs. We formally introduce labeling rules in this problem here.
\begin{definition}[Labeling rules]
    Given an unlabeled product pair $(\boldsymbol{x}_a, \boldsymbol{x}_b)$ , a labeling rule $r(\cdot)$ maps it into an label space: $r(\boldsymbol{x}_a, \boldsymbol{x}_b) \rightarrow y \in \mathcal{Y} \cup\{0\}$. Here
    $\mathcal{Y} = \{1, -1\}$ is the original label set for the task, and $\{0\}$ is a special label indicating $(\boldsymbol{x}_a, \boldsymbol{x}_b)$ is unmatchable by $r(\cdot)$. We also perform conjunction operation over atomic rules, which yields higher-order rules to capture the composable patterns.
\end{definition}

\paragraph{Problem Formulation.}
We have a unlabeled dataset $\mathcal{D}_u$ and a weakly labeled dataset $\mathcal{D}_l$, where $\mathcal{D}_l$ is generated from user behavior data thus could be noisy and incomplete. Our goal is to iteratively discover labeling rules from $\mathcal{D}_l$ to improve the weakly supervised model performance. In each iteration $t$, the high-quality rules $\mathcal{R}_t$ are used to create new weakly labeled instances $\mathcal{D}_t$ from $\mathcal{D}_u$. Using the augmented weakly labeled dataset $\mathcal{D}_l \cup \mathcal{D}_t$, we train a model $m_t: (\mathcal{X}_a, \mathcal{X}_b) \rightarrow \mathcal{Y}$. We aim to obtain a final model $f_{\theta}(\cdot): (\mathcal{X}_a, \mathcal{X}_b) \rightarrow \mathcal{Y}$ from the preceding models $\{m_t\}^T_{t=1}$ and provide interpretability by the generated rules $\{\mathcal{R}_t\}^T_{t=1}$.

%% file: 04-method.tex
\section{Methodology}
In practice, the clean data for compatible products prediction are often unavailable, and we rely on user behavior data to generate the weakly labeled dataset $\mathcal{D}_l$. Considering the noise and incompleteness of $\mathcal{D}_l$, training a machine learning model directly on $\mathcal{D}_l$ will yield suboptimal results. 
To this end, we design a boosting-style rule proposal framework to target the hard instances iteratively and propose rules adaptively. For the rule representation, we leverage information from structured product attributes and unstructured product descriptions. Such a multi-view rule generation will improve the coverage of rules while maintaining a high-precision matching.

\subsection{Adaptive Rule Proposal via Boosting}
\begin{figure}
    \centering
    \includegraphics[scale=0.275]{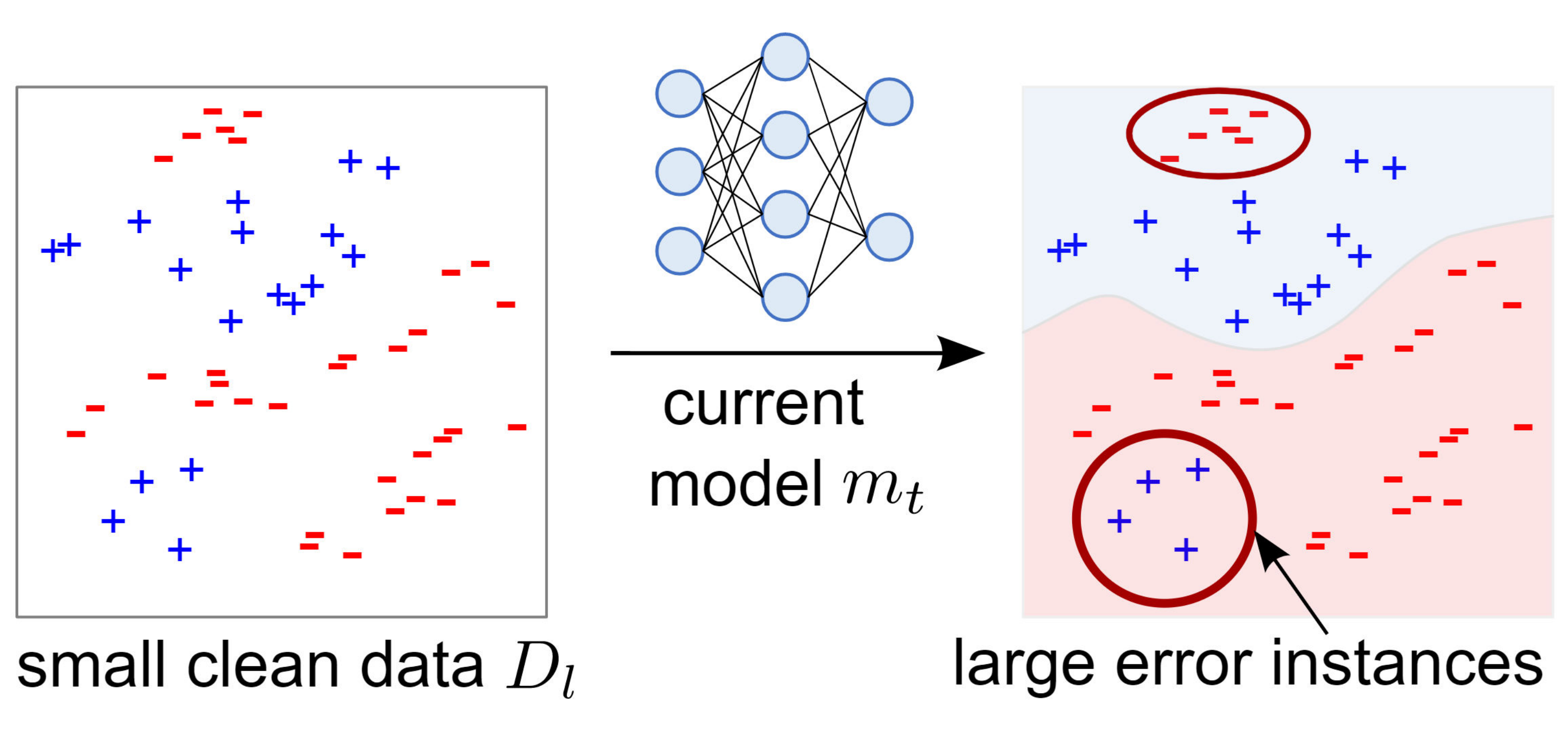}
    \caption{The boosting strategy used in \ours~ for adaptive rule proposal. The current model $m_t$ identifies the large error instances from the labeled dataset.}
    \label{fig:large_error}
\end{figure}
We adaptively target the candidate rule generation on hard instances via a boosting-style \cite{hastie2009multi} strategy. The intuition of the adaptive rule proposal is two-fold: First, the co-purchase data includes false positive samples for compatibility prediction because the frequently co-purchased product pairs are not necessarily compatible products. We aim to summarize compatibility rules from such a weak source and implement a more precise matching on the unlabeled data to improve the quality of the training set. Second, the model could be weak in certain feature regimes, we thus iteratively check such regimes by updating the data weights and propose candidate rules accordingly.

As Figure~\ref{fig:large_error} shows, the adaptive rule discovery is enabled by identifying hard instances on the labeled dataset $\mathcal{D}_l$. Through the identified hard instances where the model tends to make cumulative mistakes during the iterative learning, the discovered rules can complement the current rule set $\mathcal{R}$ and suppress the noise in the initial weak source by adding more high-quality rule-matched data, so the next model $m_{t+1}$ trained on the refined weakly labeled dataset can be improved adaptively.

Specifically, we start from the weights initialization of the instances in $\mathcal{D}_l$ as
\begin{equation}
    w_i = 1/|\mathcal{D}_l|, i = 1,2,\cdots,|\mathcal{D}_l|.
\end{equation}
During the
iterative model learning process,
each $w_i$ is updated as the model's weighted loss on instance
$\boldsymbol{x}_i \in \mathcal{D}_l$.
In iteration $t\in \{1,\cdots, T\}$, we first compute a weighted error
rate $\operatorname{err}_t$ on $\mathcal{D}_l$ by
\begin{equation}
  \operatorname{err}_t=\sum_{i=1}^{|\mathcal{D}_l|} w_{i} \mathbb{I}\left(y_{i} \neq m_t\left(\boldsymbol{x}_{i}\right)\right) / \sum_{i=1}^{|\mathcal{D}_l|} w_{i},
\end{equation}
to measure the performance of the current model $m_t$. 
Then we update the weights by
\begin{equation}
  w_{i} \leftarrow w_{i} \cdot e^{\alpha_{t} \mathbb{I}\left(y_{i} \neq m_t\left(\boldsymbol{x}_i\right)\right)}, i=1,2, \ldots, |\mathcal{D}_l|, \label{eq_weights}
\end{equation}
where the $\alpha_t$ is the coefficient of the model $m_t$:
\begin{equation}
  \label{eq:alpha}
  \alpha_t=\log \frac{1-\operatorname{err}_t}{\operatorname{err}_t}.
\end{equation}
We will use such $\alpha_t$ for later model ensemble.
In the above procedure, Equation~\ref{eq_weights} assigns a larger weight $w_i$ to sample $\boldsymbol{x}_i$ if the ensemble model consistently makes mistakes on $\boldsymbol{x}_i$
during the iterative learning, and the data points received higher weights are treated as large-error instances.
As aforementioned, a large error can be caused by the noise in the initial weakly labeled dataset. Meanwhile, when newly rule-matched instances are added into the training set, the novel local feature regimes may not  be well learned by the current model $m_t$. The weights guide
the subsequent rule proposal to target on the top-$n$ instances with the highest
$w_i$, $\mathcal{X}_e=\{\boldsymbol{x}_{j}\}_{j=1}^n$. 
In this way, we adaptively boost the model by 1) augmenting the weakly labeled dataset via iteratively adding more reliable rule-matched data, and 2) targeting the weak regimes of the current model to propose complementary rules to the current rule set.






\subsection{Multi-View Rule Generation}
\begin{table*}[t]
    \centering
    \input{table/cpt_rule}
    \caption{The examples of decision tree-based rules. The rules typically consist of two product attributes, one from the anchor product and another from the recommendation product. For example, "\emph{Battery Power Type = Battery Power Type}" indicates a categorical equivalence for two attributes with the same name; "\emph{Maximum Bulb Wattage >= Wattage}" indicates a numerical range for two different attributes. We also present a case where only one product attribute is considered. As the rule example 2 of Bathroom shows, when the "\emph{Faucet Not Included}" attribute is true for vanities, we do not match it with other faucets.}
    \label{tab:tree_rule}
    \vspace{-5mm}
\end{table*}
\ours~generates rules from two views: 1) the structured product attributes and
2) the unstructured product descriptions. Product attributes provide a better
alignment between product pairs being judged for compatibility. When the
product attributes are filled relatively completely (\textit{i.e.,} with just
a few missing values), it is ideal to apply attribute-based rules to obtain a
precise match of compatible product pairs. However, the product attributes are
of high-cardinality --- If the unlabeled instances miss some product
attributes, it will make rule matching difficult. 
We thus further leverage product description to generate \emph{fallback rules}. 
This view complements product attributes by providing text-format descriptions, where we use the \emph{sparse} product attributes to construct prompt templates in the expectation of obtaining supplementary information from product descriptions via prompting pre-trained language models. 
Although the single use of prompt-based rules may be noisy to capture the
compatibility,  it can serve as a fallback to improve the rule coverage when
the attribute-based rules are unmatched. 

\subsubsection{Rule Discovery from Structural Attributes with Decision Trees} 
To handle the categorical attributes and numerical attributes in the structural representation of products, we first design three prototypes to capture the compatibility: 
1) \emph{exact matched categorical values}, 2) \emph{within-range numerical values} 3) stand-alone attribute. Table \ref{tab:tree_rule} shows concrete examples of these rules. To express such relation among product attributes, we naturally construct a concatenated input $\boldsymbol{x}$ from the pairwise anchor products $\boldsymbol{x}_a$ and recommendation products $\boldsymbol{x}_b$:
\begin{equation}
    \boldsymbol{x} = [\boldsymbol{x}_a \oplus \boldsymbol{x}_b \oplus (\boldsymbol{x}_a^{\prime} - \boldsymbol{x}_b^{\prime}) ],
    \label{x_rep}
\end{equation}
where $\boldsymbol{x}_a^{\prime}, \boldsymbol{x}_b^{\prime}$ are the shared features of two products, \textit{i.e.}, the product attributes names in $\boldsymbol{x}_a^{\prime}, \boldsymbol{x}_b^{\prime}$ are the same.

We first base the decision tree generation on the large-error instances identified by the updated data weights:
\begin{equation}
    \mathcal{D}_t = \{\boldsymbol{x}_j\}_{j=1}^n, \quad \text{s.t.} \quad \{w_j\}_{j=1}^n = \text{Top-n}~w_i.
\end{equation}
To select features as the atomic unit of rules, we compute the permutation importance of each feature. We denote the $i$-th feature of the input $\boldsymbol{x}$ as $f_i$, and we have a evaluation metric $\phi(m_t, \mathcal{D}_l)$ to measure the performance of model $m_t$ on dataset $\mathcal{D}_l$. 
For each $k$ in $1,\cdots,K$, where $K$ is the repeat times, we first compute the $k$-th score using the same evaluation metric:
\begin{equation}
    \phi_{k,i} = \phi(m_t, \mathcal{D}_{t, i^{\prime}}).
\end{equation}
Here $\mathcal{D}_{t, i^{\prime}}$ indicates the dataset $\mathcal{D}_t$ with the $i$-th column  randomly shuffled. Then the permutation importance of feature $f_i$ can be computed by:
\begin{equation}
    \mu_i = \phi(m_t, \mathcal{D}_t) - \frac{1}{K}\sum_{k=1}^K \phi_{k,i}.
\end{equation}
Such permutation importance indicates whether a feature should be selected as the atomic unit of the rules, and we select the top-$b$ features according to the permutation importance to form the rules. 
The selected candidate rules will be evaluated in the following stage. For example, for the features belonging to $(\boldsymbol{x}_a^{\prime} - \boldsymbol{x}_b^{\prime})$, we examine if the categorical values equal to $0$, which indicates an exact match as one of the rule prototypes. For the numerical values, we output an inequality sign during rule-level annotation. We introduce details in Section \ref{sec:rule_anno}.

Though low-bias, the generated decision trees could be highly variant on the small size of identified error instances. Thus the candidate rules consisted of important nodes on the trees need to be further evaluated by humans. Moreover, the selected features may correspond to some sparse product attributes. When a quite number of features are filled with placeholder values, it is hard to summarize insightful information from the product attributes. Even we generate rules from such features, only a few unlabeled instances can be matched with these rules. To this end, we introduce the following prompt-based rules to handle the sparse but important product attributes. \\

\subsubsection{Rule Discovery from Unstructured Descriptions with Prompting.}  
Although decision tree-based rules are capable of matching unlabeled instances by checking product attributes, the high cardinality of attributes makes the rule generation and rule matching process sometimes nontrivial. In this case, we further leverage product descriptions to propose prompt-based rules as fallback rules. 

Specifically, we construct rule templates based on the sparse product
attributes selected from the previous step. For example, we denote the
selected sparse feature as $f^{\prime}$, the product names of product A and B
are $\boldsymbol{n}_a, \boldsymbol{n}_b$, and the product description of them
are $\textbf{Desc}_a, \textbf{Desc}_b$. Then the rule
prompt can be constructed from a template, where the original product descriptions and
 the selected product attribute are connected by some template words. We show a concrete example in Table~\ref{tab:prompt_rule_example}.
\begin{table*}[]
    \centering
    \begin{tabular}{rp{15cm}}\hline
         Product & Product Description\\\hline
         Power tool & The Milwaukee M18 FUEL 1/2 in. Hammer Drill is the industry's most powerful brushless battery powered drill, delivering up to 60\% more power. \\
         Battery & The Milwaukee M18 REDLITHIUM XC 5.0 Ah Battery Pack delivers up to 2.5X more runtime, 20\% more power and 2X more recharges than standard lithium-ion batteries.\\\hline
         Template & $\boldsymbol{n}_a$: $\textbf{Desc}_a$. $\boldsymbol{n}_b$: $\textbf{Desc}_b$. The $\boldsymbol{n}_a$ is (not) compatible with the $\boldsymbol{n}_b$ because their $\boldsymbol{f}^{\prime}$ are \texttt{[MASK]}.\\
         Prompt & \textbf{Power tool}: The Milwaukee M18 FUEL 1/2 in. Hammer Drill is the industry's most powerful brushless battery powered drill, delivering up to 60\% more power. \textbf{Battery}: The Milwaukee M18 REDLITHIUM XC 5.0 Ah Battery Pack delivers up to 2.5X more runtime, 20\% more power and 2X more recharges than standard lithium-ion batteries. The \textbf{power tool} is \emph{compatible} with the \textbf{battery} because their \underline{brand names} are \texttt{[same]}. \\\hline
    \end{tabular}
    \caption{The example of prompt-based rules. In the prompt, we underline the selected feature $\boldsymbol{f}^{\prime}$ and use bold font to highlight the product name $\boldsymbol{n}_a$,  $\boldsymbol{n}_b$. The token at the end is the prediction of the \texttt{[MASK]} token given by the PLM. The italic word "\emph{compatible}" is given by the weak label of this product pair.}
    \vspace{-5mm}
    \label{tab:prompt_rule_example}
\end{table*}

In this way, we provide the context from the product description and prompt the PLMs to fill the \texttt{[MASK]} token. 
By filling the masked slot in the prompt, PLMs propose candidate
keyword-based rules for topic classification. Different from the rules extracted
from surface patterns of the corpus (\textit{e.g.}, $n$-gram rules), such a prompt-based rule proposal can generate words that do not appear in the original inputs---this capability is important to model
generalization.

Given the selected sparse feature $f^{\prime}$ and a large-error instance $\boldsymbol{x}_{e} \in \mathcal{D}_t$ such that $\boldsymbol{x}_{e}$ miss a original value for the feature $f^{\prime}$.
we first convert it into a prompt
$\boldsymbol{x}_{p}$ by the rule template above,
which consists of product descriptions of the original input pair and the compatibility label, as well as a \texttt{[MASK]} token to be predicted by a PLM $\mathcal{M}$. 
To complete the rule, we feed $\boldsymbol{x}_{p}$ to $\mathcal{M}$ to obtain the probability distribution of the \texttt{[MASK]} token over the vocabulary $\mathcal{V}$:
\begin{align}
 \setlength{\abovedisplayskip}{0.05pt}
\setlength{\belowdisplayskip}{0.05pt}
\begin{small}
  \label{prob_task}
  p(\texttt{MASK} = \hat{\mathbf{v}}\mid\boldsymbol{x}_{p})
  = \frac{\exp \left(\hat{\mathbf{v}} \cdot \mathcal{M}(\boldsymbol{x}_{p})\right)}{\sum\limits_{\mathbf{v} \in \mathcal{V}} \exp \left(\mathbf{v} \cdot \mathcal{M}(\boldsymbol{x}_{p}\right))},
 \end{small}
\end{align}
where $\mathcal{M}(\cdot)$ denotes the output vector of $\mathcal{M}$.  We use the prediction with highest $p(\texttt{MASK} = \hat{\mathbf{v}} \mid \boldsymbol{x}_{p})$
to form the candidate rules. By filling the rule template of $\boldsymbol{x}_{e}$ with its prompt prediction, we obtain a candidate prompt-based rule. Integrated with the proceeding decision tree-based rules, we obtain the rule set in iteration $t$, denoted as $\mathcal{R}_t = \{r_{j}\}_{j=1}^{b}$. Here the number $b$ is same as the number of selected features. In other words, for each selected important feature, we generate a rule either by the decision tree or the prompt.

\subsection{Rule Annotation and Rule Matching}
\label{sec:rule_anno}
The candidate rule set $\mathcal{R}_t$ is just one step away from the rule matching stage. Note that the selected features just specify the column on the product attribute table, so we need a concrete value for each feature to complete the rule.
\ours~thus presents $\mathcal{R}_t$ to humans for evaluation, which can yield a set of positive rules. 
Specifically, for each candidate rule $r_j\in R_t$, humans can make a decision from the following options:
1) \emph{Exact match}: This operation is set for the features belonging to $(\boldsymbol{x}_a^{\prime} - \boldsymbol{x}_b^{\prime})$. For example, the base code of chandelier and lighting bulb should be an exact match to be compatible.
2) \emph{Range}: This operation is set for the features in $(\boldsymbol{x}_a^{\prime} \oplus \boldsymbol{x}_b^{\prime})$. For example, the bulb wattage should be lower than the maximum wattage.
3) \emph{Contain}: This operation is set for the features in $(\boldsymbol{x}_a^{\prime} \oplus \boldsymbol{x}_b^{\prime})$. It does not require a pairwise relation between the product pair, but requires the selected  attribute of one product to have its original value instead of a filled placeholder.
4) \emph{Abstain}: Human annotators can reject this rule due to its poor quality.

After the rule-level annotation, the accepted rules will be used for weak label generation via rule matching. We compute a matching score weighted by the feature importance to decide if an unlabeled product pair should be assigned a weak positive label. The matching score is computed in two steps. First, for the decision tree-base rules, we only need to check the hard match of the specific product attributes. If matched, the score will be accumulated with the corresponding feature importance:
\begin{equation}
    s_{t,j}^d = \mu_j \mathbb{I}(r_j == f_j^u),
\end{equation}
where $f_j^u$ is the $j$-th feature of the unlabeled instance.

However, as we mentioned above, some attributes could be sparse among the corpus, thus resulting in difficulty for matching. As a fallback, the prompt-based rules are exactly proposed for this situation by checking the semantic similarity in the product descriptions.
The integration of the both views of rules is enabled by the computation of prompt matching score. We first feed
$\tau(\mathbf{x}_t^u)$ into the prompting model (Equation \ref{prob_task}) and
use the to obtain the prompt of instance
$\mathbf{x}_t^u$. Then we compute the prompt matching score:
\begin{equation}
  s_{t,j}^p = \mu_j \frac{\mathbf{e}_t^u\cdot\mathbf{e}_j^r}{
  \|\mathbf{e}_t^u\|\cdot\|\mathbf{e}_j^r\|},
\end{equation}
where $\mathbf{e}_t^u$ is the prompt embedding of $\mathbf{x}_t^u$ and $\mathbf{e}_j^r$ is the rule embedding of $r_j$, both embeddings are obtained from a PLM encoder.
Next, we compute the final matching score by merging decision tree-based rules and the prompt-based rules:
\begin{equation}
     s_{t,j} = \sum_j^b ( s_{t,j}^d + s_{t,j}^p) = \sum_j^b \mu_j(\mathbb{I}(r_j == f_j^u) + \frac{\mathbf{e}_t^u\cdot\mathbf{e}_j^r}{
  \|\mathbf{e}_t^u\|\cdot\|\mathbf{e}_j^r\|}).
\end{equation}

\subsection{Model Learning and Ensemble}
\begin{figure}
    \centering
    \includegraphics[scale=0.5]{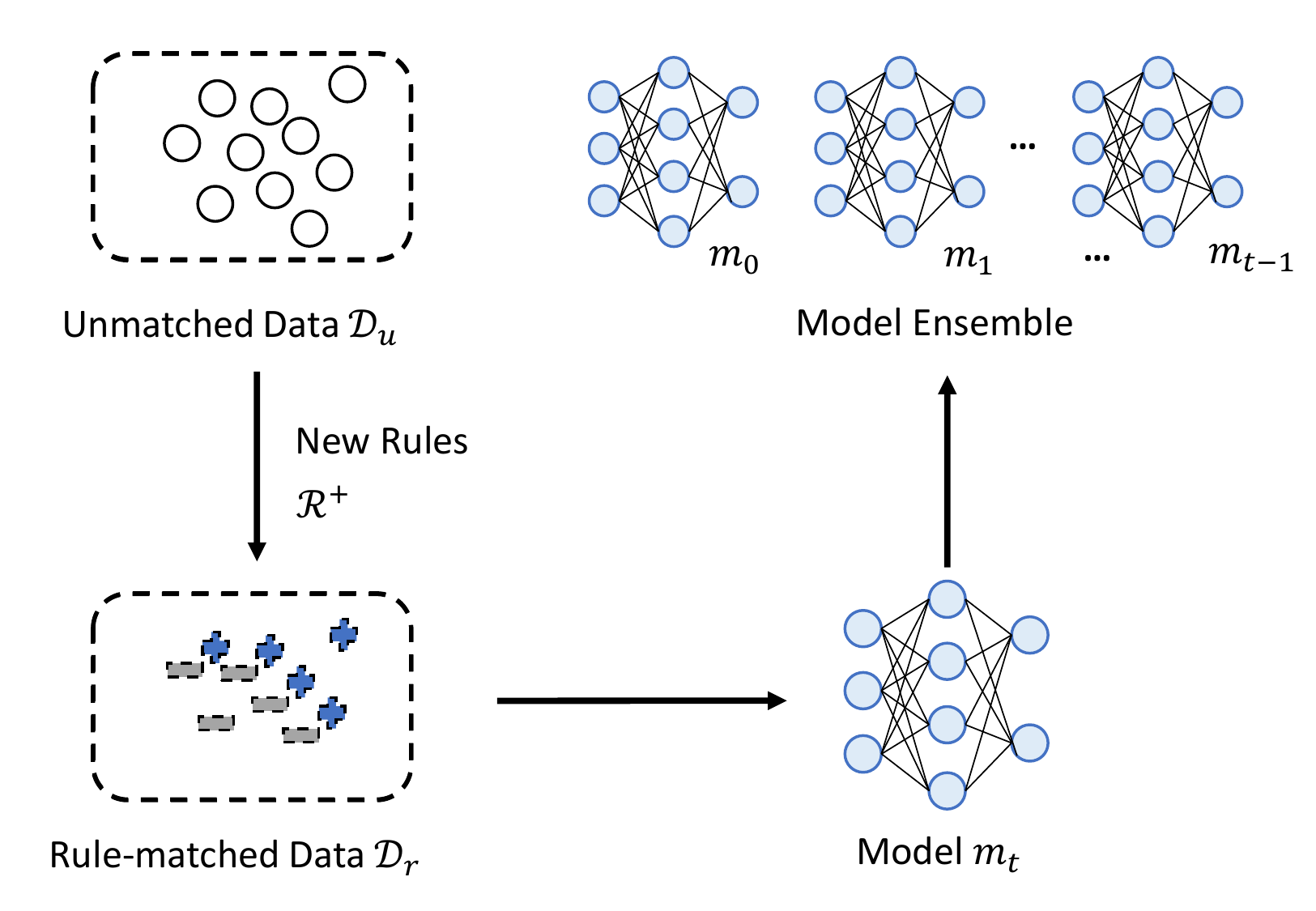}
    \caption{Rule matching and model ensemble.}
    \label{fig:ensemble}
    \vspace{-8mm}
\end{figure}
Figure~\ref{fig:ensemble} shows how we proceed from rule matching to the model learning and ensemble. Specifically, in iteration $t$, with the new rule-matched data $\mathcal{D}_r$ integrated, we obtain an enlarged weakly labeled dataset $\mathcal{D}_t = \mathcal{D}_{t-1} \cup \mathcal{D}_r$. We fit a weak model $m_t$ on $\mathcal{D}_t$ by optimizing:
\begin{equation}
 \setlength{\abovedisplayskip}{0.1pt}
\setlength{\belowdisplayskip}{0.1pt}
  \min _{\theta} \frac{1}{\left|\mathcal{D}_{t}\right|} \sum\limits_{\left(\boldsymbol{x}_i,\hat{y}_{i}\right) \in\mathcal{D}_{t}} \ell_{\operatorname{CE}}\left(m_t(\boldsymbol{x}_i), \hat{y}_{i}\right),
\end{equation}
where $\hat{y}_i$ is the weak label for instance $\boldsymbol{x}_i$,  and $\ell_{\operatorname{CE}}$ is the cross entropy loss.

Finally, we incorporate the newly trained weak model into the ensemble model. The final model is a weighted ensemble of the proceeding models:
\begin{equation}
 \setlength{\abovedisplayskip}{0.15pt}
\setlength{\belowdisplayskip}{0.15pt}
  f_{\theta}(\cdot) = \sum_t^n \alpha_t m_t,
  \label{eq:ensemble}
\end{equation}
where a weak model $m_t$ with a low error rate $\operatorname{err}_t$ will be assigned a higher coefficient $\alpha_t$ according to Equation \ref{eq:alpha}.

%% file: table/cpt_rule.tex
{
\begin{tabular}{ccccc}\hline
     Dataset & Anchor Category & Rec. Category & Rule Example 1 & Rule Example 2\\\hline
     Lighting & Lighting Fixture & Light Bulbs & Compatible Bulb Type = Lighting Technology &Maximum Bulb Wattage >= Wattage  \\
     Appliance & Refrigerator & Water Filter & Walter Filter Replacement Model\# = Model\# & Brand Name = Brand Compatibility\\
     Bathroom& Vanities & Faucets& Color Family = Color Family & Faucet Not Included \\
     Tools& Power Tools & Batteries& Battery Power Type = Battery Power Type & Voltage <= Voltage\\\hline
\end{tabular}
}

%% file: 05-exp.tex
\section{Experiments}
\subsection{Experiment Setup}

\subsubsection{Data}
We create four benchmarks for compatible product prediction from real-world
data obtained from \emph{homedepot.com}. Each dataset consists of 5k product
pairs sampled from a specific anchor product category and a specific
recommendation product category. To provide a comprehensive evaluation, the
product categories covered in the four datasets are selected as diverse as
possible, including 1) \textbf{Lighting}: \emph{Lighting Fixture and Light
  Bulbs}; 2) \textbf{Appliance}: \emph{Refrigerators and Water Filters}; 3)
\textbf{Bathroom}: \emph{Vanities and Faucets}; 4) \textbf{Tools}: \emph
{Power Tools and Batteries}. To avoid the expensive human annotation process,
we use co-purchase data as the weak source to sample positive instances. Given
the specific category pair, we can query the compatible products from the
highly-frequent co-purchase product pairs. For the negative instances, we
randomly sample product pairs from the same category, but use the co-purchase
data to filter out those possible compatible products. Although such a process
can introduce noise, our annotators with domain expertise The Home Depot, the
co-purchase data can be regarded as a reliable weak source.

\subsubsection{Evaluation Protocol}
For the 5k weakly labeled dataset generated from co-purchase data, we use the ratio $7$:$1.5$:$1.5$ to split train, validation and test set. Besides, we have a hold-out 5k dataset for querying labels in AL settings or matching unlabeled instances by \ours. To make a convincing test set, we use a higher co-purchase score to distinguish the data quality between the test set and the training set, thus making the test set derived from user behavior data, though weakly supervised, still a reasonable proxy for measuring compatibility. We use 
classification accuracy as the evaluation metric.



\subsubsection{Baselines} We compare our method with the following
baselines: 
\textbf{MLP}: A multi-layer perceptron classifier with 2 hidden layers.
\textbf{DPClass-F}~\cite{shang2016dpclass}: It implements pattern-based classification by mining discriminative patterns from the prefix paths in random forest. It selects patterns with the biggest improvement of the training classification accuracy.
\textbf{DPClass-L}~\cite{shang2016dpclass}: A variant of DPClass. Its pattern selection is based on an L1 regularization term. 
\textbf{Self-training}~\cite{lee2013pseudo}: It is the vanilla self-training method that generates pseudo labels for unlabeled data.
\textbf{Entropy}-based active learning~\cite{holub2008entropy,yu2021atm}: It is an uncertainty-based method that acquires annotations on samples with the highest predictive entropy. 
\textbf{CAL}~\cite{margatina2021active} is the most recent AL method for pre-trained LMs. It calculates the uncertainty of each sample based on the KL divergence between the prediction of itself and its neighbors' prediction.\\

\subsubsection{Implementation Details}
We use a MLP with 2 hidden layers as the classifier in all experiments. For the optimizer, we use AdamW \cite{loshchilov2018decoupled} and choose learning rate from $\{2 \times 10^{-4}, 1\times 10^{-4}, 5\times 10^{-5}\}$. For the decision tree, we search the depth between $[3, 10]$ with an interval of $1$. For the PLM, we choose RoBERTa-base~\cite{liu2019roberta} for rule discovery with prompting. We keep the number of iterations as 10 for all the tasks and the annotation budget $b=10$ in each iteration. For iterative baselines, we query 300 data points in each iteration. We set the repetition time $K=10$ for the feature importance calculation.

\subsection{Main Result}
\begin{table}[]
    \centering
    \input{table/main_res}
    \caption{Main results on four real-world datasets.}
    \label{tab:main}
    \vspace{-8mm}
\end{table}

Table~\ref{tab:main} reports the performance of \ours~ and the baselines on
the 4 benchmarks. The results demonstrate \ours~ consistently outperforms baselines on all the four datasets. \ours~ significantly outperforms its underlying MLP, the improvement on four datasets are $7.33\%, 5.85\%, 5.55\%, 5.22\%$, respectively. Compared to the strongest interactive learning baseline CAL, \ours~ still holds a $0.93\%$ lead averaged on the four datasets. The performance improvement to self-training baseline ranges from $1.00\% \sim 2.34\%$, which demonstrates the advantage of our adaptive rule discovery for enhancing WSL over the vanilla weak label generation paradigm. 

\begin{figure}
     \centering
     \begin{subfigure}[b]{0.23\textwidth}
         \centering
         \includegraphics[width=\textwidth]{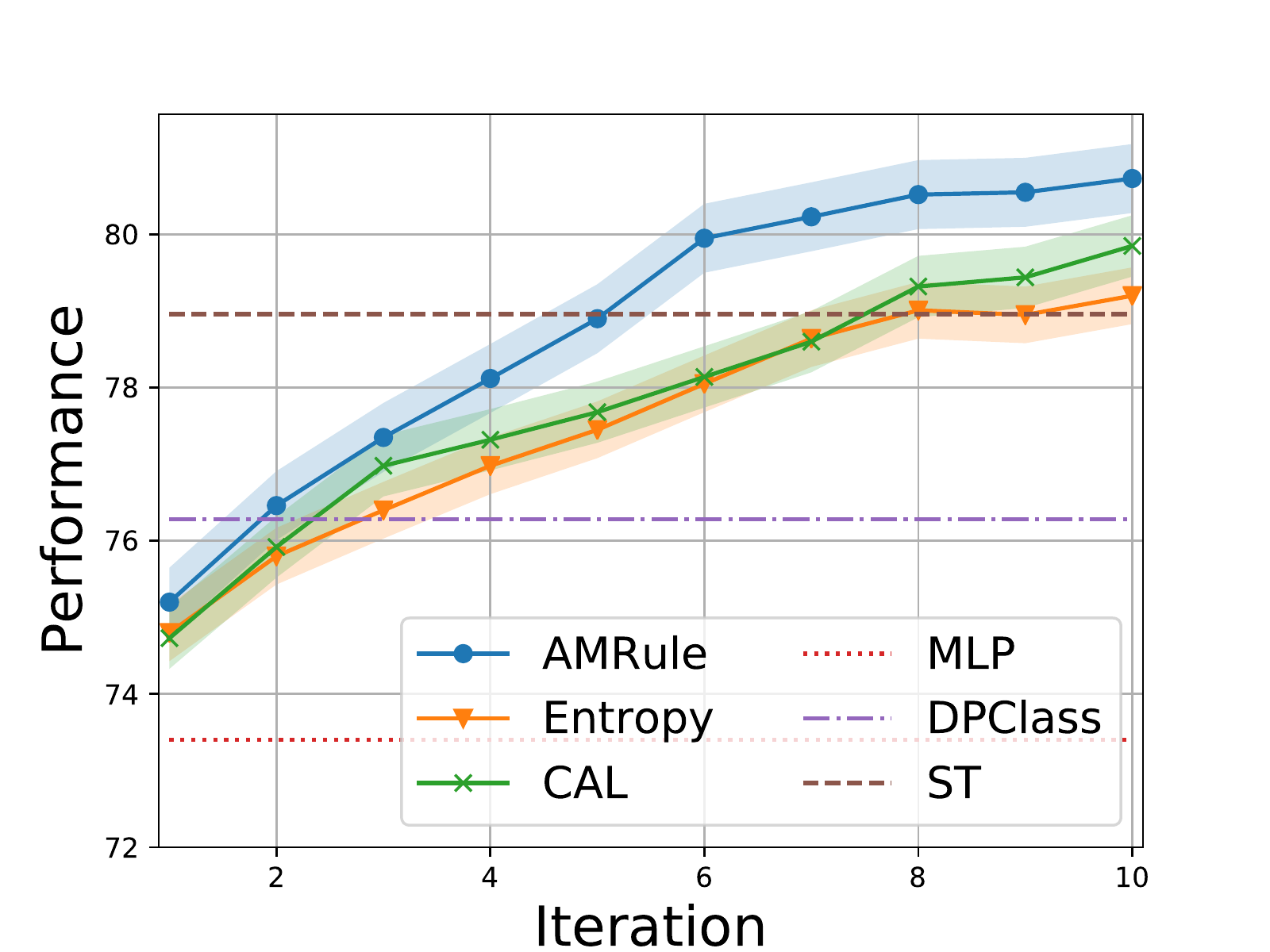}
         \caption{Lighting}
         \label{fig:y equals x}
     \end{subfigure}
     \hfill
     \begin{subfigure}[b]{0.23\textwidth}
         \centering
         \includegraphics[width=\textwidth]{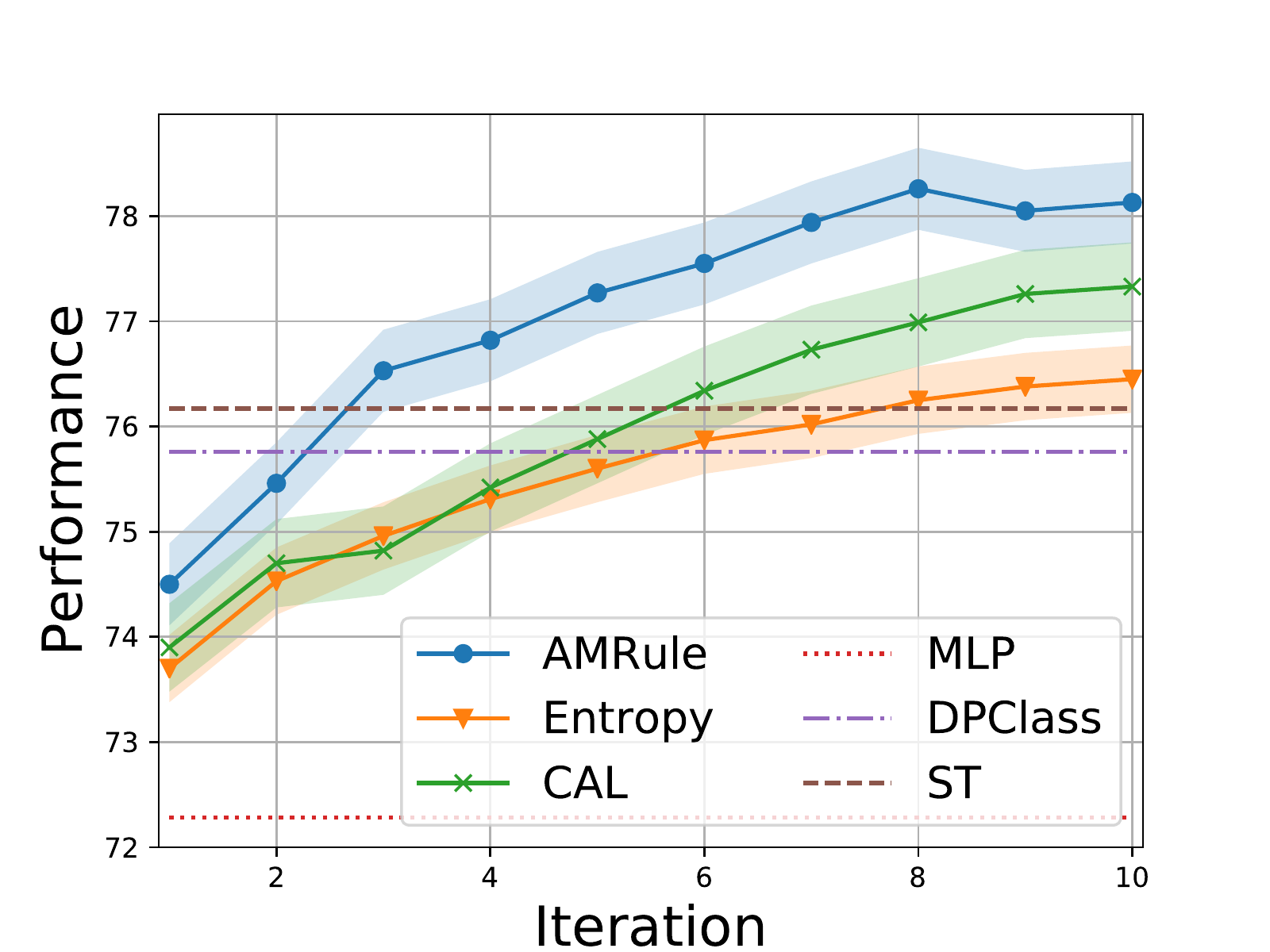}
         \caption{Appliance}
         \label{fig:y equals x}
     \end{subfigure}
     \hfill
     \begin{subfigure}[b]{0.23\textwidth}
         \centering
         \includegraphics[width=\textwidth]{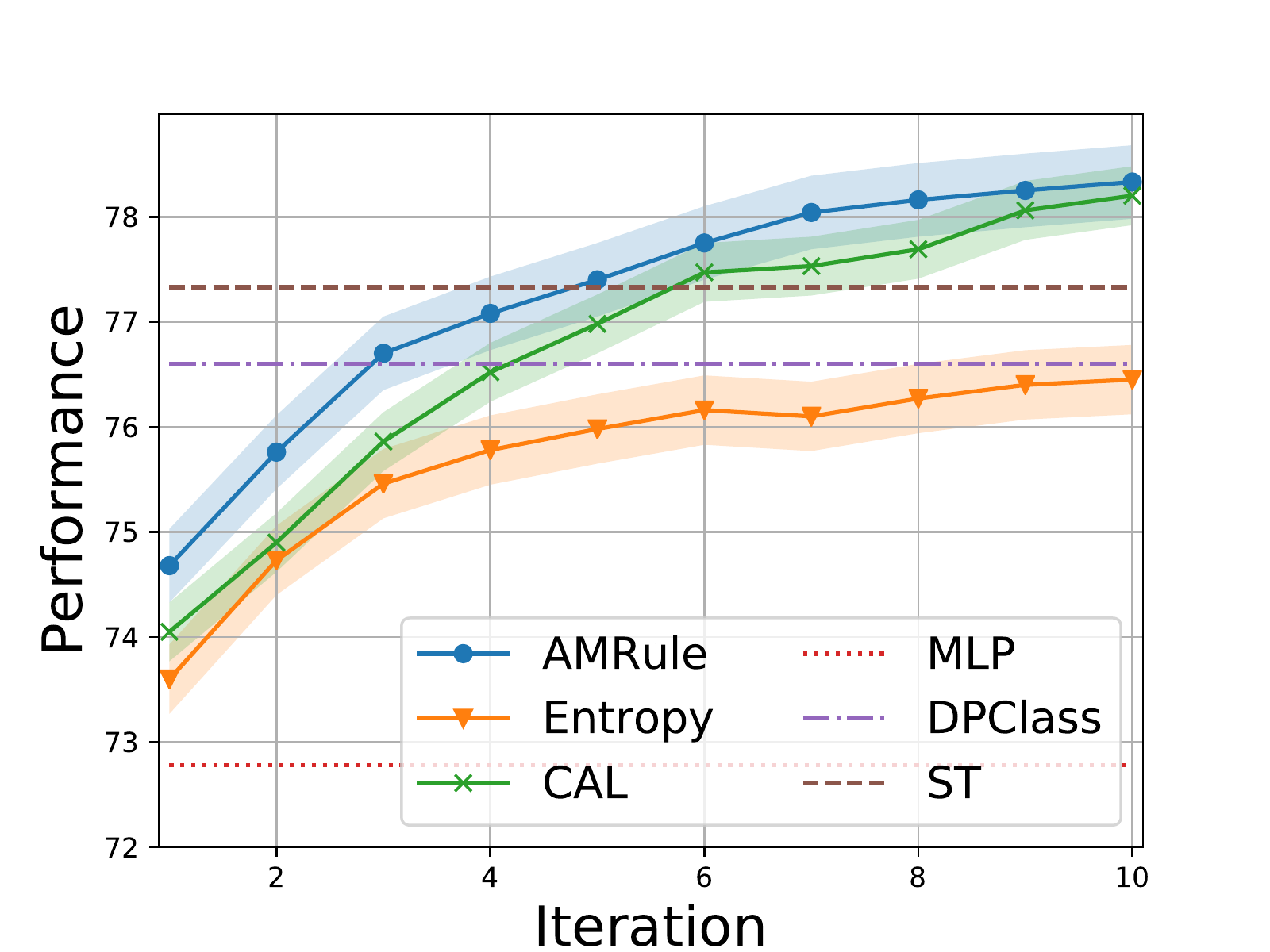}
         \caption{Bathroom}
         \label{fig:three sin x}
     \end{subfigure}
     \hfill
     \begin{subfigure}[b]{0.23\textwidth}
         \centering
         \includegraphics[width=\textwidth]{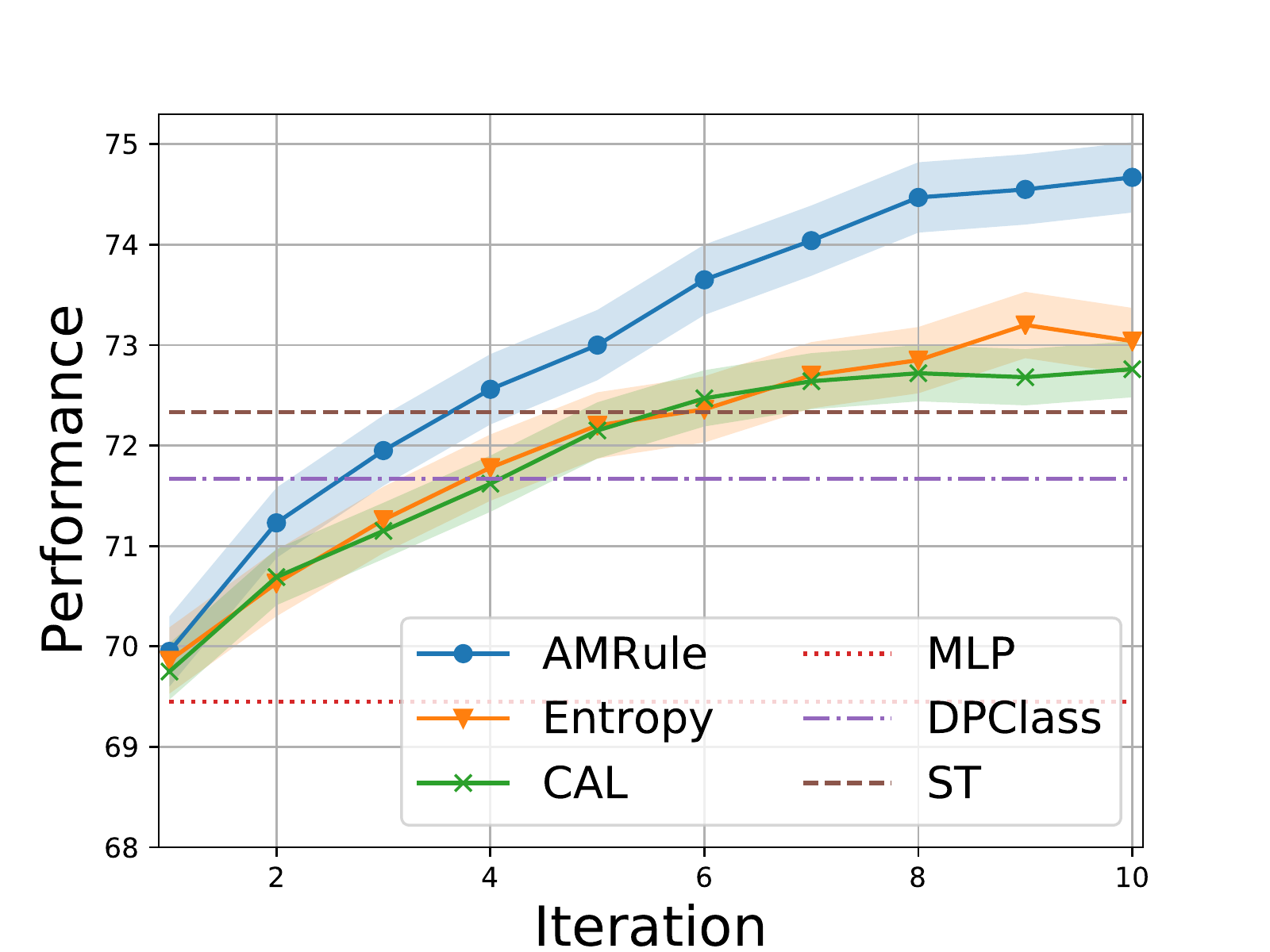}
         \caption{Tools}
         \label{fig:five over x}
     \end{subfigure}
        \caption{The performance in terms of different number of iterations. The horizontal lines belong to the static methods, the rising curves belong to iterative methods, where they actively query new labeled data or discover rules to generate more weakly labeled data.}
        \label{fig:iter_res}
\end{figure}

We present the iterative performance in  Figure~\ref{fig:iter_res}, compared with the baselines that also feature iterative learning. We observed \ours~outperforms the other iterative methods consistently in each iteration. It is worth noting that \ours~also reduces much annotation cost compared to the iterative baselines, which adopt the instance-level annotation. For the rule-level annotation in \ours, it takes each annotator nearly 100 seconds to annotate 10 candidate rules, which is just $2\%$ of the instance-level annotation time for 500 unlabeled points in each iteration.  

We notice that the performance improvement varies by product category. For the Bathroom dataset, the baseline CAL performs closely to \ours, while the performance on other datasets presents a large margin between \ours~ and baselines. It is because compatibility rules vary by product category. For example, the Bathroom dataset takes vanities and faucets as the anchor and recommendation product category. During the rule discovery process, we find the candidate rules focus on a few product attributes, which describe colors and dimensions. In contrast, when the rule discovery provides a diverse candidate rule set where different product attributes are covered, more unlabeled instances will be covered during the rule matching stage, thus the end model presents a stronger performance over the baselines. 

\subsection{Analysis of the Multi-view Rules}
\begin{figure}
    \centering
    \includegraphics[scale=0.4]{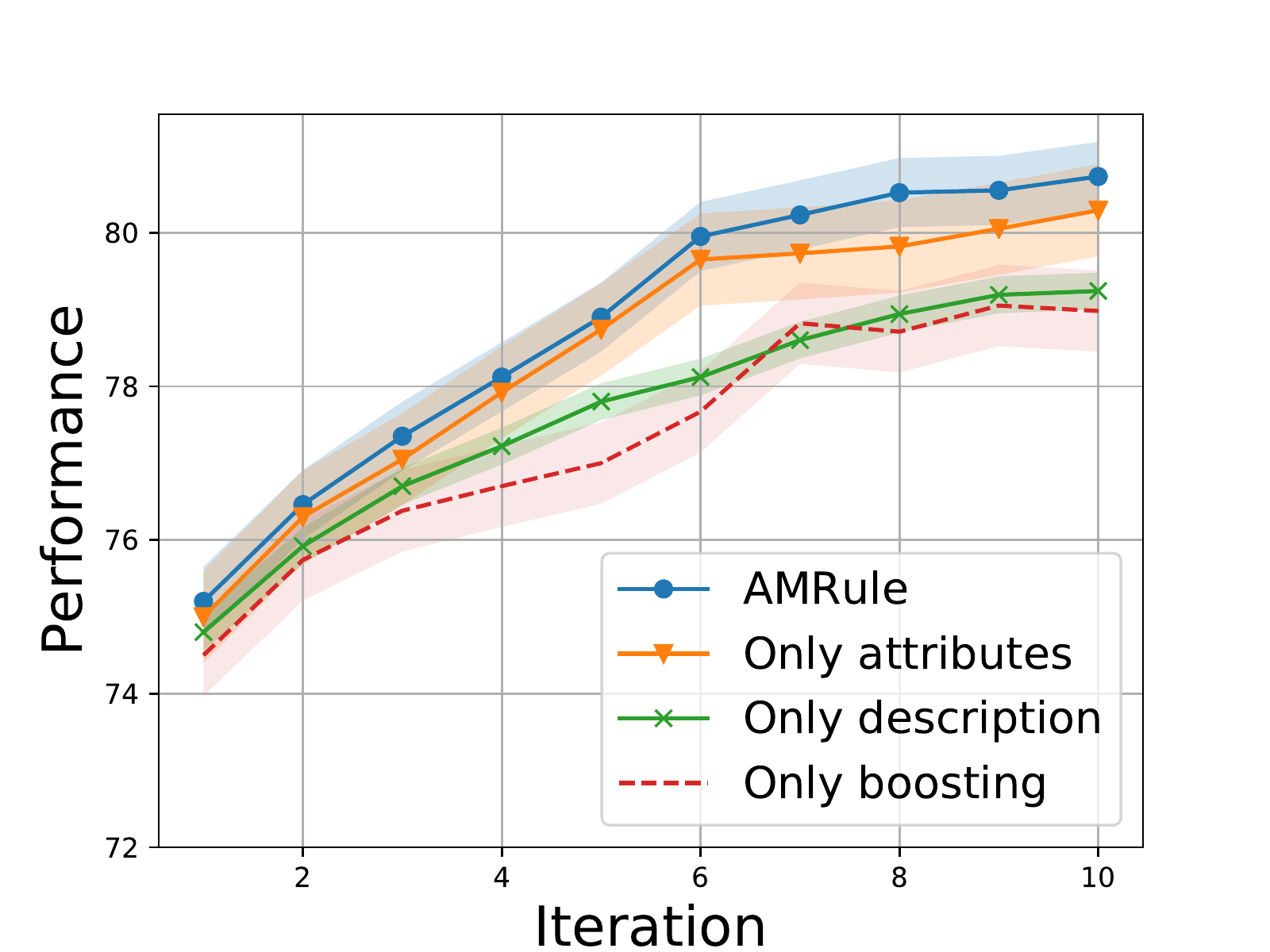}
    \caption{A closer look at the effect of multi-view rules. \ours~ is the complete method where we discover rules from both the structured attributes and the unstructured descriptions. `Only attributes' and `Only description' denote we only consider a single view of rule. `Only boosting' means the implementation does not discover any rules but only iteratively train weak models and ensemble model.}
    \label{fig:multiview}
    \vspace{-5mm}
\end{figure}

In this set of experiments, we study the effect of different views of rules in \ours. Figure~\ref{fig:multiview} verifies the advantage of the multi-view rule design, when both views of rules are considered, the model presents the best prediction capability. When we implement a single-view rule discovery, we found the rules generated from structured attributes work better than the rules generated from unstructured descriptions. 
It can be explained by the high relevance between the product attributes and product compatibility. The structured attributes can precisely match with unlabeled product pairs, because it excludes those incompatible products via hard filtering.  However, as shown in Figure~\ref{fig:multiview}, only proposing rules based on the structured attributes will fall behind the multi-view approach significantly after several iterations. When the selected features become increasingly sparse with the iteration increases, the attribute-only  rules can hardly match enough unlabeled instances to further improve the model. At this time, the product descriptions can serve as a complementary source for rule discovery. It undertakes the task of rule discovery from the sparse attributes by prompting PLMs to mine the semantic hint in the unstructured descriptions. In this way, it supplements the attribute view by matching unlabeled instances in a soft manner, where we use prompts for the descriptions of unlabeled instances and then compute a similarity score. The results demonstrate both views of rules do complement each other and yield the best performance when integrated.

\subsection{Effect of different components}
\begin{figure}
    \centering
    \includegraphics[scale=0.4]{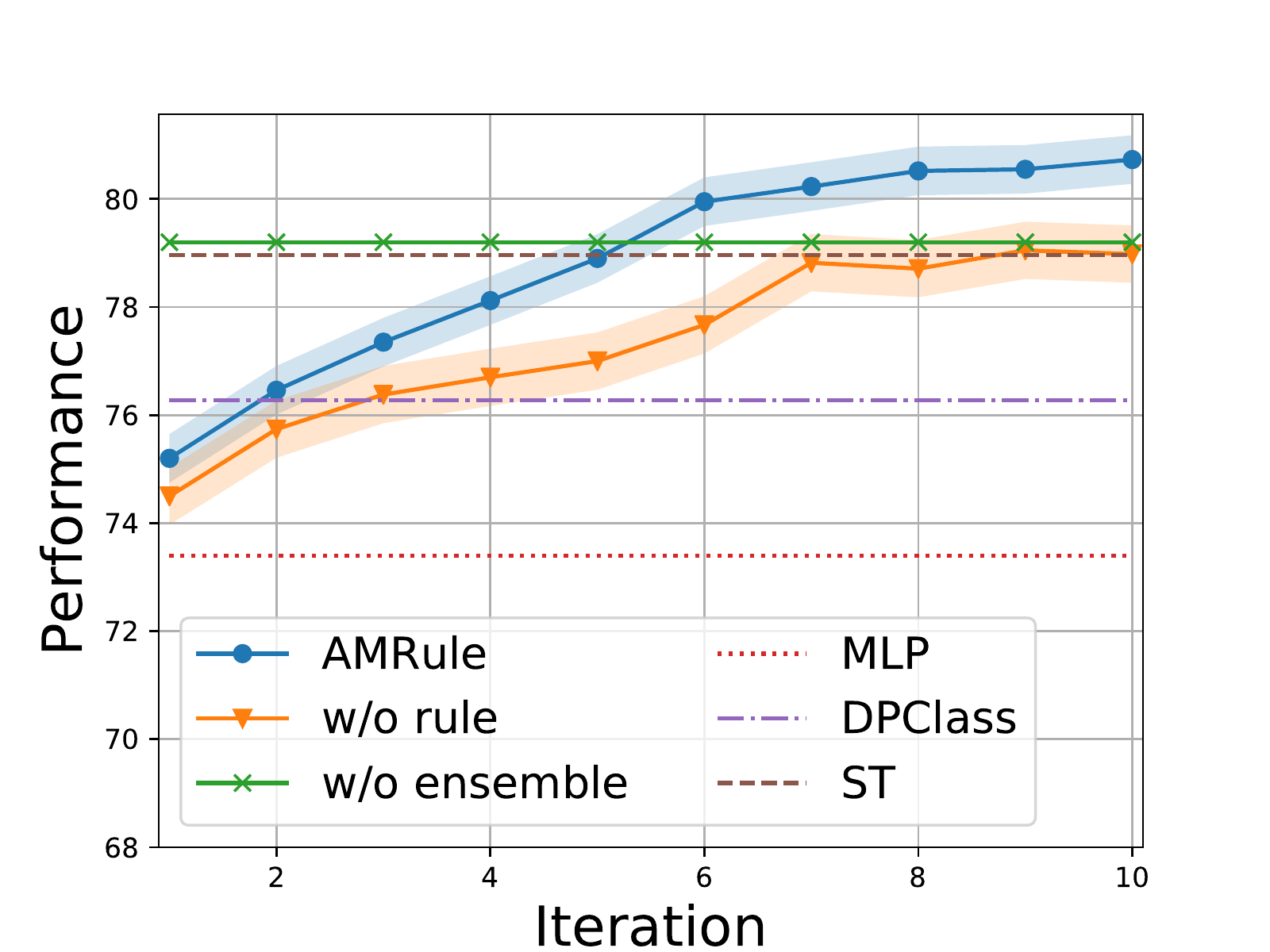}
    \caption{Ablation study on Lighting dataset. 'w/o rule' indicates we do not propose rules during the iterative process; 'w/o' ensemble indicates we use up the annotation budget in the first iteration to discover rules and do not implement the adaptive rule discovery process.}
    \label{fig:ablation}
    \vspace{-5mm}
\end{figure}

Figure~\ref{fig:ablation} presents the ablation study to verify the effect of different components in \ours, our findings are as follows:

1) Adaptive rule discovery by targeting hard instances is effective.  In Figure~\ref{fig:ablation},
the "w/o ensemble" setting shows the adaptive strategy contributes to the final performance. 
Specifically, we fix the annotation budget $b$ but
discover candidate rules from large-error samples in one iteration. 
We compare \ours~ with this "w/o ensemble" curve and observed a $1.53\%$ performance gain, which comes from that the complete method iteratively identifies the hard instances and proposes rules based on these instances accordingly.
Such an adaptive process helps the model to check its weaknesses and refine them by discovering new rules, thus yielding more effective rules than static rule discovery under the same annotation budget.

2) Ensemble alone without new rule discovery is not as effective. For
the "w/o rule" variant, we do not propose new rules, but ensemble multiple
self-trained weak classifiers instead. The final performance drops
significantly under this setting by $1.75\%$. It demonstrates the newly
proposed rules provide complementary weak supervision to the model. Although
simply ensembling multiple weak classifiers also helps WSL, it is not as
effective as training multiple complementary weak models as in \ours.\\



\subsection{Case Study}
We use a case study to illustrate the rule discovery process. Here the anchor category is power tools and the recommendation category is batteries. In iteration $t$, \ours~ first select the features based on the permutation importance. The selected features include [\emph{battery power type}, \emph{brand name}, $\cdots$, \emph{number of total batteries included}]. For the battery type which belongs to $ (\boldsymbol{x}_a^{\prime} - \boldsymbol{x}_b^{\prime})$ in Equation \ref{x_rep}, the human annotator can implement the operation "Exact match" indicating this product attributes should be associated with a condition of categorical equivalence. For example, the battery power type of the anchor and recommendation products are both "Lithium Ion", and it is a reasonable compatibility hint. For the next candidate \emph{brand name}, it is a sparse attribute under this category pair, so we leverage product descriptions to generate a prompt-based rule. As Table \ref{tab:prompt_rule_example} shows, we can get a rule by prompting a PLM. For the last candidate \emph{number of total batteries included} which is a stand-alone and irrelevant feature, the human annotator can choose the "Abstain" operation to reject this rule.  

%% file: table/main_res.tex
{
\begin{tabular}{ccccc}\toprule
    \textbf{Method} & \textbf{Lighting} & \textbf{Appliance} & \textbf{Bathroom} & \textbf{Tools}  \\\midrule
    MLP & 73.40 & 72.28&  72.78& 69.45\\
    DPClass-F & 76.28& 75.58 & 75.86& 71.67\\
    DPClass-L & 75.92 & 75.76 & 76.60&71.42 \\\hline
    Self-training & 78.96 & 76.17& 77.33 & 72.33 \\
    Entropy AL &79.20& 76.45& 77.16&73.04  \\
    CAL & 79.85& 77.33& 78.20 & 72.76 \\\hline
    \ours~ & 80.73  & 78.13&  78.33 & 74.67 \\\bottomrule
\end{tabular}
}

%% file: 06-conclusion.tex
\section{Conclusion}
To solve the compatible products prediction task, we propose an adaptive rule discovery framework \ours~ to predict compatibility from weakly labeled data. We leverage user behavior data to construct a weakly supervised setting, which avoids cumbersome annotation engineering. From the heterogeneous nature of the product representation, we design the multi-view rules to incorporate both structured product attributes and unstructured product descriptions. Through the decision tree-based and prompt-based rule generation, \ours~ discovers multi-view rules that complement each other to improve the weakly supervised model. Moreover, the boosting strategy enables identifying hard instances for rule discovery, so the new rules iteratively refine where the model accumulates mispredictions. Experiments on real-world benchmarks demonstrate \ours~ significantly improves weakly-supervised compatible products prediction and provides interpretable predictions by discovering compatibility rules.